\documentclass{article}

\PassOptionsToPackage{numbers, compress}{natbib}
\usepackage[preprint]{neurips_2026}

\usepackage[utf8]{inputenc} % allow utf-8 input
\usepackage[T1]{fontenc}    % use 8-bit T1 fonts
\usepackage{hyperref}       % hyperlinks
\usepackage{url}            % simple URL typesetting
\usepackage{booktabs}       % professional-quality tables
\usepackage{amsmath}        % math environments
\usepackage{amsfonts}       % blackboard math symbols
\usepackage{nicefrac}       % compact symbols for 1/2, etc.
\usepackage{microtype}      % microtypography
\usepackage{xcolor}         % colors
\usepackage[table]{xcolor}
\usepackage{comment}

\usepackage{pifont}

\usepackage{booktabs}
\usepackage{multirow}
\usepackage{amssymb}
\usepackage{graphicx}
\usepackage{placeins}

\title{P-JEPA: Procedural Video Representation Learning via Joint Embedding Predictive Architecture}

\author{%
  Felix Tristram \\
  Technical University of Munich\\
  Munich Center for Machine Learning\\
  \texttt{felix.tristram@tum.de} \\
  \And
  Stefano Gasperini \\
  Technical University of Munich\\
  Munich Center for Machine Learning\\
  \texttt{stefano.gasperini@tum.de} \\
  \And
  Benjamin Killeen \\
  Technical University of Munich\\
  Munich Center for Machine Learning\\
  \texttt{benjamin.killeen@tum.de} \\
  \And
  Marcel Walch \\
  Carl Zeiss AG\\
  \texttt{marcel.walch@zeiss.com} \\
  \And
  Christian Benz \\
  Carl Zeiss AG\\
  \texttt{christian.benz@zeiss.com} \\
  \And
  Nassir Navab \\
  Technical University of Munich\\
  Munich Center for Machine Learning\\
  \texttt{nassir.navab@tum.de} \\
  \And
  Ghazal Ghazaei \\
  Carl Zeiss AG\\
  \texttt{ghazal.ghazaei@zeiss.com} \\
}

\begin{document}

\maketitle

\begin{abstract}
%Real-world tasks are characterized by long-range hierarchical structures, in which individual actions are meaningful only within their broader context. Therefore, long-horizon video understanding requires representations that preserve local motion while modeling how actions evolve over minutes of activity. Current VideoLLMs can ingest long videos through sparse sampling or memory, but often delegate temporal reasoning to language models over highly compressed visual evidence. Conversely, strong video encoders such as I3D, TSM, and V-JEPA produce useful short-range features, but do not learn reusable take-level temporal structure alone. We propose Procedural-JEPA (P-JEPA), a self-supervised temporal representation learner for long video feature streams. P-JEPA trains a shallow JEPA model over complete takes: a student observes masked feature sequences, a predictor reconstructs latent teacher targets, and an EMA teacher provides stable targets without action labels, narrations, or procedural graphs. This objective encourages representations to encode action order, procedural progress, and context-dependent action meaning across long video horizons. Across EgoExo4D, EgoProceL, and Assembly101, P-JEPA improves linear separability and streaming inference, and it consistently strengthens temporal action segmentation models when used as their input representation, demonstrating the effectiveness of P-JEPA as a self-supervised temporal learner.

The increasing maturity of embodied AI platforms has driven a growing interest in procedural video representation learning to support intelligent assistance systems for complex, multi-step tasks.
% Procedural videos capture complex, multi-step tasks that 
% Understanding human actions is crucial for embodied AI systems to imitate them or for AI assistants to provide helpful advice.
Leveraging large-scale latent predictive training, video foundation models capture video dynamics, enabling downstream tasks such as activity understanding, spatiotemporal localization, and predictive control.
However, procedural videos include actions with long-range dependencies that these models do not support, due to the quadratic complexity of self-attention.
Distinct actions, for example, may be visually similar despite appearing at different points in the procedure, such as turning the stove on versus off.
Here, we propose a backbone-agnostic approach that learns long-duration video representations by reducing the problem to a dense, frame-aligned action space and predicting pooled masked latent vectors.
This approach allows our Procedural Joint Embedding Predictive Architecture (P-JEPA) to ingest videos over 30 minutes long, enabling effective long-form understanding of procedural steps.
We evaluate P-JEPA using features extracted with VJEPA2.1, TSM, and I3D over the EgoExo4D, EgoProceL, and Assembly101 datasets, finding that it consistently improves linear separability, streaming inference, and temporal action segmentation performance, achieving state-of-the-art results on EgoExo4D fine-grained action classification while using an order of magnitude fewer parameters than LLM-based methods and running in real time.
\end{abstract}

\section{Introduction}

Many real-world activities are procedural: they unfold as ordered sequences of actions whose meaning depends on the surrounding task context. Cooking, repair, assembly, and skilled manipulation all require understanding not only what is visible in a short clip, but also which steps have already occurred, which state changes have been induced, and where the current action lies within the broader procedure~\cite{plizzari2025omnia}. Across procedural videos, visually similar motions can correspond to different steps depending on temporal context, making it hard to distinguish them from visual cues alone: reaching for a tool, placing a component, or manipulating a stove knob may indicate preparation, assembly, adjustment, or completion depending on the preceding actions. 
Procedural video understanding, therefore, requires representations that combine fine-grained local motion with long-range temporal structure. 
%This need also extends to embodied AI, where agents must interpret ongoing procedures over time to operate reliably in long-horizon tasks.

Existing spatio-temporal video backbones provide strong local video features and have achieved impressive performance on short-term action understanding \cite{carreira2017quo,lin2019tsm,arnab2021vivit,bertasius2021space, liu2022video,fan2021multiscale,neimark2021video}. Self-supervised approaches extend this paradigm by training latent predictive models on millions of video hours \cite{bardes2024revisiting, vjepa2, mur2026vjepa2_1}.
However, all of these models are typically trained and evaluated on short clips of a few seconds. Directly applying them to minute-long or take-level videos is computationally expensive, since the number of frames or patch tokens grows rapidly with video length and self-attention scales quadratically with token count. Thus, long procedural video understanding requires a mechanism to model dependencies across much longer action sequences without incurring prohibitive sequence growth.

Recent VideoLLMs address this challenge by connecting visual encoders to large language models and extending the video context through sparse frame sampling, token compression, or external memory \cite{song2024moviechat,song2025moviechat+,he2024ma,tan2024koala, ren2024timechat,chen2024videollm,shu2025video}. While these systems are effective for open-ended video question answering and retrieval-style tasks, recent diagnostic benchmarks suggest that long context alone does not guarantee genuine temporal reasoning. Single-frame baselines have been shown to be surprisingly competitive on several long-video QA benchmarks \cite{ranasinghe2024understanding,plizzari2025omnia,wu2024longvideobench}, indicating that many examples can be solved from static cues or sparse retrieval alone. For questions that require reasoning about event sequences, however, most VideoLLMs remain substantially weaker. 
%Benchmarks such as EgoTempo~\cite{plizzari2025omnia} and LongVideoBench~\cite{wu2024longvideobench} make this limitation explicit: while many perception-level questions can be answered from static visual cues, VideoLLMs remain substantially weaker on relation-level questions that require temporal order, event sequences or before/after relations. 
Other dedicated temporal evaluations show that even strong image-video multimodal LLMs \cite{chen2024internvl,wang2024qwen2} struggle with precise temporal grounding and temporal reasoning~\cite{liu2024tempcompass,videomme, li2024temporal}. These findings motivate us to explore video understanding without language models. 

Temporal modeling without LLMs is typically done via learning dense frame-wise action labels over untrimmed videos, where a model has to assign a class to each frame. For this temporal action segmentation task, the field has developed strong models: Temporal convolutional networks, multi-stage refinement, and recent transformer-based architectures can model action order and long-range temporal context effectively when trained with dense annotations~\cite{farha2019ms,li2020mstcn,singhania2023c2f,yi2021asformer, behrmann2022unified,bahrami2023much,lu2024fact}. 
So far, however, investigation into the input to these models has been limited, with some form of short-term feature extractor providing dense frame-wise features \cite{lin2019tsm, carreira2017quo}.
Additionally, these models are typically supervised predictors for a fixed dataset and label space, which allows them to learn temporal structure for a particular action taxonomy. This limits their use when labels are sparse, when task taxonomies differ across datasets, or when one wants a reusable representation for diverse procedural videos.

One way of learning such re-usable procedural representations is to align the visual evidence from instructional videos with narrations, procedural knowledge graphs, text-derived step structure, or hierarchical descriptions \cite{zhong2023learning,zhou2023procedure,peirone2025hiero}. 
Some works go further and learn key steps and procedural order through matching of recurring visual patterns across demonstrations alone \cite{elhamifar2019unsupervised,bansal2024united}. 
%Other methods study procedural structure with weaker or no manual supervision by discovering key steps and their ordering from multiple videos of the same task, often through alignment of recurring visual patterns across demonstrations~\cite{elhamifar2019unsupervised,bansal2024united}. 
%These approaches highlight the importance of procedural structure, but often rely on additional language supervision, procedural knowledge graphs, repeated demonstrations, or objectives centered on step discovery rather than dense frame-aligned representation learning.
Similar to these unsupervised works we propose to learn procedural representations solely from data but do so for dense frame-aligned features. 

We achieve this by reducing the data from a noisy RGB distribution to a frame-aligned action space and training a Joint Embedding Predictive Architecture (JEPA) to discover procedural structure in the training data distribution. Leveraging features from pretrained video encoders such as I3D~\cite{carreira2017quo}, TSM~\cite{lin2019tsm}, and V-JEPA~2.1~\cite{mur2026vjepa2_1} our Procedural-JEPA (P-JEPA) model applies a masked latent prediction objective inspired by JEPA-style video representation learning~\cite{bardes2024revisiting,vjepa2, mur2026vjepa2_1} but applies it to video-takes over 30 minute long. 
Despite receiving no action labels, narrations, transcripts or procedural graphs, this simple objective induces representations that capture temporal order, procedural progress, and context-dependent action meaning. 
This is not obvious a priori: masked prediction over pretrained visual features could instead learn only appearance similarity, short-term smoothness, or dataset-specific co-occurrence. Our experiments show that take-level latent prediction reorganizes the feature space around procedural structure, making segments more linearly separable, improving streaming inference, and producing feature trajectories that align more strongly with within-video temporal progress.

%P-JEPA operates on frame-wise feature streams from pretrained video backbones such as I3D~\cite{carreira2017quo}, TSM~\cite{lin2019tsm}, and V-JEPA~2.1~\cite{mur2026vjepa2_1}. Leveraging these features from long take-level sequences, P-JEPA applies a masked latent prediction objective inspired by the JEPA-style video representation learning~\cite{bardes2024revisiting,vjepa2, mur2026vjepa2_1} to complete procedures: a student encoder observes visible tokens, a predictor fills masked positions, and an EMA teacher provides latent targets.

%We evaluate P-JEPA across multiple procedural video datasets, feature backbones, and downstream temporal evaluations.

Our main contributions can be summarized as follows:
\begin{itemize}
    \item We introduce P-JEPA, a take-level predictive representation learning framework for long procedural videos.
    \item We demonstrate that masked latent prediction over dense feature streams induces temporally-aware representations without explicit supervision.
    %through narrations, transcripts, procedural knowledge graphs, repeated demonstrations, or semantic action labels.
    \item We evaluate P-JEPA across I3D, TSM, and V-JEPA~2.1 feature streams and show improvements under linear, streaming, and temporal action segmentation evaluations, with state-of-the-art fine-grained action classification on EgoExo4D~\cite{grauman2024ego} and strong gains on EgoProceL \cite{bansal2022egoprocel} and Assembly101 \cite{sener2022assembly101}.
\end{itemize}
\section{Related Work}

\textbf{Video encoders and clip-level representation learning.}
Video understanding has long relied on architectures that extract local
spatio-temporal features from short clips. I3D~\cite{carreira2017quo} extends
2D convolutional networks to video by inflating filters into 3D, jointly
modeling appearance and motion. TSM~\cite{lin2019tsm} preserves the efficiency
of 2D CNNs by shifting part of the feature channels along the temporal
dimension. Transformer-based models, including ViViT~\cite{arnab2021vivit},
TimeSformer~\cite{bertasius2021space}, Video Swin~\cite{liu2022video}, and
MViT~\cite{fan2021multiscale}, further improve short-clip recognition through
spatio-temporal attention, hierarchy, or multi-scale token representations.

Self-supervised video encoders learn such features from unlabeled clips through
reconstruction, contrastive objectives, or latent prediction. Masked
reconstruction methods learn from missing visual content~\cite{tong2022videomae},
while predictive joint-embedding methods avoid pixel-level generation and
instead predict latent representations~\cite{bardes2024revisiting,vjepa2,
mur2026vjepa2_1}. These models
provide strong local representations, but are typically trained and evaluated
over short temporal windows. P-JEPA instead uses such encoders as feature
extractors and learns a take-level representation over complete procedures.
%\vspace{-0.3cm}

\textbf{Long-video reasoning with vision-language models.}
Recent VideoLLMs extend multimodal LLMs to longer videos by sparsely sampling
frames, compressing visual tokens, or maintaining external memory
banks~\cite{song2024moviechat,song2025moviechat+,he2024ma,tan2024koala,
ren2024timechat,chen2024videollm,shu2025video}. These systems are effective for
video question answering and retrieval-style tasks, but diagnostic studies show
that text-only, question-only, or single-frame baselines can perform
surprisingly well on many long-video QA benchmarks
~\cite{ranasinghe2024understanding,plizzari2025omnia,wu2024longvideobench}.
Related analyses further indicate that temporal reasoning can be a bottleneck
of the language model itself, rather than only a limitation of visual
features~\cite{li2024temporal}. VL-JEPA~\cite{chen2025vl} also adopts a
predictive joint-embedding formulation, but predicts language-target embeddings
for vision-language tasks. In contrast, P-JEPA uses visual latent targets and
learns dense procedural representations for temporal action understanding,
without language-conditioned supervision.
%\vspace{-0.3cm}

\textbf{Temporal action segmentation.}
Temporal action segmentation is the supervised setting most closely related to
our evaluation, as it requires dense action predictions over untrimmed videos.
Early neural approaches established temporal convolutional networks as an
effective way to model dense action sequences from pretrained frame-wise
features. Multi-stage TCNs use dilated temporal convolutions, iterative
prediction refinement, and smoothing losses to reduce over-segmentation and
improve temporal consistency~\cite{farha2019ms,li2020mstcn}. Later
architectures extend this principle with a stronger multi-scale structure, for
example, by predicting from coarse to fine temporal resolutions
~\cite{singhania2023c2f}, or by combining local temporal modeling with
longer-range context. Transformer-based methods further introduce locality,
hierarchy, windowed attention, sparse long-range attention, or explicit
segment-level representations to make dense long-video prediction practical
~\cite{yi2021asformer,behrmann2022unified,bahrami2023much,lu2024fact}.
P-JEPA is complementary to these supervised sequence predictors: it learns a procedural
representation that can be used as input to these models. 
%\vspace{-0.3cm}

\textbf{Procedure-aware and long-range video representation learning.}
Several works explicitly study procedural structure in videos, aiming to model
not only which action is happening, but also how actions form steps, tasks, and
higher-level procedures. One line learns from instructional videos paired with
narrations or transcripts, using language to infer step concepts, temporal
ordering, or task-level structure~\cite{zhong2023learning,
zhou2023procedure}. Procedure-aware pretraining methods such as
PaPrika~\cite{zhou2023procedure} further use procedural knowledge graphs or
text-derived pseudo-labels to supervise representations for task, step, and
next-step understanding. More recent hierarchical approaches, such as
HiERO~\cite{peirone2025hiero}, enrich clip representations with higher-level
activity context.

Other methods study long-range or procedural structure with weaker supervision.
Unsupervised procedure-learning methods discover key steps and their ordering
from multiple videos of the same task by aligning recurring visual patterns
across demonstrations~\cite{elhamifar2019unsupervised,bansal2024united}.
Self-supervised hierarchical methods separate local visual encoding from
higher-level event modeling, combine short-term consistency with global video
structure, predict future representations, or align clips with narrations and
whole-video summaries~\cite{xiao2022hierarchical,qing2022learning,
mounir2023streamer,ashutosh2023hiervl}. Related work on state-change
counterfactuals further emphasizes that the before/after structure is important for
procedural egocentric understanding~\cite{kung2025changed}. P-JEPA shares the
goal of learning procedure-aware representations, but does so from individual
complete takes, without narrations, transcripts, procedural graphs, repeated
demonstrations, or semantic action labels during representation learning.
\section{Method}

\subsection{Feature Extraction}
\label{sec:features}

Directly modeling RGB frames over minute-long videos is challenging for transformer-based architectures: the dense frame-level input quickly becomes computationally prohibitive due to quadratic self-attention scaling, and the noisy low-level image space makes it difficult to capture long-term dependencies. We therefore use frame-wise features extracted with VJEPA2.1 \cite{mur2026vjepa2_1}, TSM \cite{lin2019tsm} or I3D \cite{carreira2017quo} as input to P-JEPA. 

More formally, for any of the aforementioned models we decompose the frame-wise features of a video into an ordered sequence of segments $s=1,\ldots,S$, where a segment is defined by the start/end time of its corresponding annotation. Each Segment $s$ then contributes
$m_s$ tokens
\[
    \mathbf{X}_s =
    \left(\mathbf{x}_{1},\ldots,\mathbf{x}_{m_s}\right),
    \qquad
    \mathbf{x}_{t}\in\mathbb{R}^{d_{\mathrm{in}}}.
\]
The take is the concatenated stream
$\mathbf{X}=\mathbf{X}_1\Vert\cdots\Vert\mathbf{X}_S$ with
$T=\sum_{s=1}^{S}m_s$ valid tokens. Each token index $i$ in $T$ has both a \textit{segment} index $c(i)$ and a \textit{within-segment} index $r(i)$ representing its frame-index within the segment.

For I3D and TSM we use existing features and do not train the models ourselves. For V-JEPA2.1 we reduce its patch-wise output to a single token per frame to align with the other architectures and allow long-video learning. We do so using two different approaches: for a simple baseline, we apply a non-learned mean pooling (\textit{mean-pool}) to the output over the spatial dimension, resulting in one token per timestep. The second approach uses a learnable cross-attention pooler, which, given the output features $\mathbf{U}\in\mathbb{R}^{n\times d_v}$ from the frozen VJEPA2.1 encoder of a fixed number of frames $n$, applies a small self-attention stack and then pools to $q$ learnable queries: 
\begin{equation}
    \mathbf{V}_n
    =
    \mathrm{CrossAttn}\!\left(\mathbf{Q},
    f_{\mathrm{self}}(\mathbf{U})\right)
    \in\mathbb{R}^{q\times d_v}.
\end{equation}
These $\mathbf{V}_n$ tokens are then classified with a small attentive classifier using a standard cross-entropy loss on the action labels provided by each dataset. After convergence, the classifier is discarded and the pooled tokens $\mathbf{V}_n$ are extracted with non-overlapping sliding windows for each segment. We term these features \textit{action-pool} and illustrate the pipeline in the supplementary. We set $q$ to match input frames, resulting in one token per frame.

For all feature extractors a learned LayerNorm--linear projection maps input tokens from $d_{\mathrm{in}}$ to the P-JEPA model width $d$ if $d \neq d_{\mathrm{in}}$.

%We use two feature sources. For EgoExo4D, a frozen V-JEPA~2.1 encoder
%\cite{vjepa2} extracts short-clip spatio-temporal tokens. Since directly
%processing all V-JEPA tokens over a multi-minute take is infeasible, we learn a
%query-based action pooler. Given V-JEPA features
%$\mathbf{U}_s\in\mathbb{R}^{n_s\times d_v}$ for clip $s$, we first apply a
%small self-attention stack and then pool with $q$ learned queries,

%A classifier on top of the pooled tokens is trained with cross-entropy on clip-level action labels. 

%For Assembly101 and EgoProceL, we also evaluate P-JEPA on dense TSM features. The Temporal Shift Module (TSM) \cite{lin2019tsm} shifts a fraction of feature channels forward and backward along the time dimension inside a 2D CNN, enabling neighboring frames to exchange information without adding parameters or FLOPs. We sample the resulting frame-level features at the dataset-specific rate and slice them by segment boundaries, producing variable-length token streams. 

\subsection{Self-Supervised Procedural Learning}
\label{sec:ssl_procedure}

\begin{figure}[t]
\vspace{-0.3cm}
\begin{center}
\includegraphics[width=0.8\linewidth]{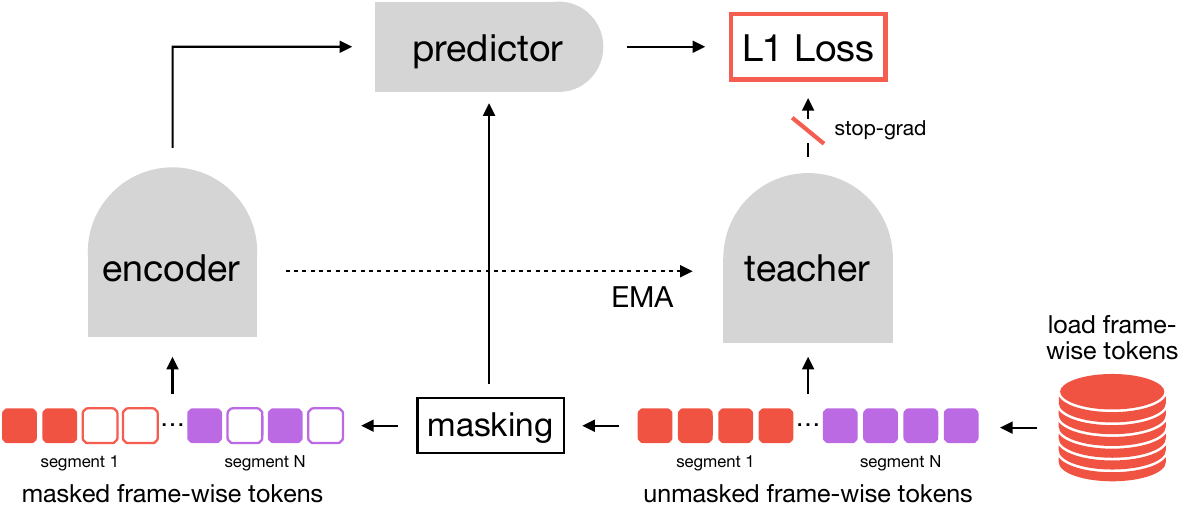}
\end{center}
\vspace{-0.2cm}
\caption{\textbf{P-JEPA}. A take-level JEPA is trained on continuous streams of
feature tokens. The student encoder sees randomly selected context tokens, the
predictor fills masked target positions, and an EMA teacher provides latent
targets for the complete valid stream.}
\label{fig:ssl_framework}
\vspace{-0.3cm}
\end{figure}

After feature extraction, each video is represented as an ordered stream of
tokens rather than raw RGB frames. This lets us move from short-window visual
recognition to the central question of procedural understanding: can a model
learn not only what a segment looks like, but where it belongs in the broader
procedure? In many tasks, the same local visual evidence can correspond to
different actions depending on what happened before. For example, reaching
toward a stove knob may mean turning the stove on, adjusting the heat, or
turning it off. Similarly, filling a pot with water is meaningful partly because
it precedes later steps such as boiling or adding ingredients. We therefore want
each token to become \emph{procedure-aware}: informed by the current segment,
the preceding sequence of events, and the temporal structure of the take.

P-JEPA learns such representations with a self-supervised predictive objective
over the full take-level token stream $\mathbf{X}$. Instead of predicting pixels
or using action labels, the model learns by hiding parts of the procedural
sequence and predicting their latent teacher representations from the visible
context. For padded batches, let $\Omega\subseteq\{1,\ldots,T\}$ be the
set of valid tokens. We sample a target set
$\mathcal{M}\subset\Omega$ uniformly at random with masking ratio $p_m$, and
define the context set $\mathcal{C}=\Omega\setminus\mathcal{M}$. The student
encoder $f_\phi$ receives only context tokens, the predictor $g_\psi$ predicts
masked positions in latent space, and the teacher encoder $f_{\bar{\phi}}$
receives all valid tokens. The teacher parameters are an exponential moving
average of the student parameters and receive no gradients.

Concretely, the predictor input is a full-length sequence containing student
features at visible positions and a learned mask token at target positions.
The loss is only applied to mask tokens $\mathcal{M}$
\begin{equation}
\mathcal{L}_{\mathrm{P\mbox{-}JEPA}}
= \frac{1}{|\mathcal{M}|}
\sum_{i\in\mathcal{M}}
\left\|
    \hat{\mathbf{z}}_i -
    \mathbf{z}_i
\right\|_1,
\qquad
\hat{\mathbf{z}}_{\mathcal{M}}
=
\left[g_\psi(\mathrm{fill}(f_\phi(\mathbf{X}_{\mathcal{C}})))\right]_{\mathcal{M}},
\quad
\mathbf{z}_{\mathcal{M}}
=
\left[f_{\bar{\phi}}(\mathbf{X}_{\Omega})\right]_{\mathcal{M}} .
\end{equation}
The prediction problem therefore forces the encoder to organize tokens according
to the regularities of the procedure itself: which events tend to occur
together, which states precede others, and how the current segment relates to
the history of the take. We illustrate the method in Figure \ref{fig:ssl_framework}.

\begin{figure}[h]
\begin{center}
\includegraphics[width=1.0\linewidth]{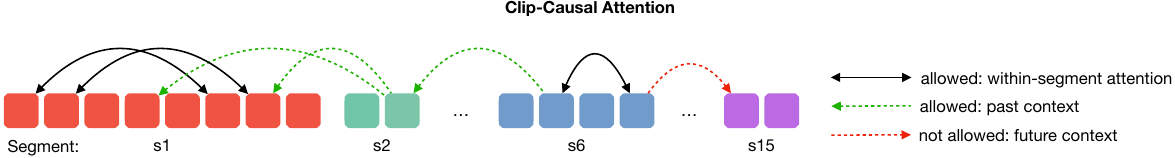}
\vspace{-0.7cm}
\end{center}
\caption{\textbf{Clip-Causal Attention}. All tokens \textit{within} a segment are allowed to attend to each other (black) and to past segments (green). Attention to future segments is blocked (red).}
\label{fig:clip_causal}
%\vspace{-0.3cm}
\end{figure}

\textbf{Clip-Causal Attention.}
A key design choice is how much temporal context each representation is allowed
to use. If attention were fully bidirectional, a token could rely on future
segments and would no longer represent an online or streaming setting. If it
were strictly token-causal, however, tokens inside the same annotated segment
could not fully exchange information, even though they describe the same local
step. We therefore propose clip-causal attention: tokens may attend freely within
their own segment and to all previous segments, but never to future segments.

More specifically, for an encoder query token $i$ and key token $j$, we allow
attention only when the key is visible and belongs to the current or a previous
segment:
\begin{equation}
    A^{\mathrm{enc}}_{ij}
    =
    \mathbb{I}\{j\in\mathcal{K}\}
    \mathbb{I}\{c(j)\le c(i)\},
    \qquad
    \mathcal{K}\in\{\mathcal{C},\Omega\},
    \qquad \mathbb{I}\{\cdot\} = \text{Indicator Function}
\end{equation}
This \emph{clip-causal} mask is fully visible within a segment but causal across segments. It lets the representation of segment $s$ use all evidence inside segment $s$ and all past segments, while preventing access to future segments. The predictor remains a standard token-causal transformer over the flattened stream, using $A^{\mathrm{pred}}_{ij}=\mathbb{I}[j\in\Omega]\mathbb{I}[j\le i]$.
%\vspace{-0.3cm}

\textbf{2D RoPE.}
The attention mask defines which tokens can exchange information, but the model
still needs to know where each token lies in the procedure. A token has two
relevant coordinates: its segment index, which captures progress through the
take, and its within-segment index, which captures local frame-wise order.
To encode both, we apply separate rotary positional embeddings along the two
coordinates $(c(i),r(i))$. For a token $\mathbf{x}_{s,t}$, each attention head produces
$\mathbf{q}^{(h)}_{s,t},\mathbf{k}^{(h)}_{s,t}\in\mathbb{R}^{d_h}$, split as
$d_h=d_S+d_T+d_0$. We rotate the first part by segment index and the second
part by local token index:
\begin{align*}
\tilde{\mathbf{q}}^{(h)}_{s,t}
&=
\big[
R_S(s)\mathbf{q}^{(h,S)}_{s,t},
\;
R_T(t)\mathbf{q}^{(h,T)}_{s,t},
\;
\mathbf{q}^{(h,0)}_{s,t}
\big], \\
\tilde{\mathbf{k}}^{(h)}_{s,t}
&=
\big[
R_S(s)\mathbf{k}^{(h,S)}_{s,t},
\;
R_T(t)\mathbf{k}^{(h,T)}_{s,t},
\;
\mathbf{k}^{(h,0)}_{s,t}
\big].
\end{align*}
The rotation matrices $R_S$ and $R_T$ use the inverse-frequency bands of RoPE
\cite{su2024roformer}; the remaining $d_0$ channels are unrotated.
%\vspace{-0.3cm}

\begin{figure}[h]
\begin{center}
\includegraphics[width=0.7\linewidth]{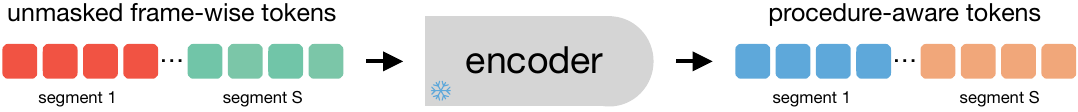}
\end{center}
\vspace{-0.8em}
\caption{During inference, only the P-JEPA encoder is used. Input tokens are
unmasked and are contextualized by current-segment and past-segment evidence.}
\label{fig:inference}
\vspace{-0.2cm}
\end{figure}

\textbf{Inference.}
At inference time we discard the predictor and teacher and use only the student encoder as a procedure encoder. It is run without masking on the full valid token stream, producing clip-causal representations that combine current-segment evidence with the history of the take. These contextualized representations are then used for all downstream probes.

\subsection{Evaluation of Representations}
\label{sec:probe}

To quantify the impact of P-JEPA's learned representations on downstream temporal and procedural understanding tasks, we follow standard representation learning practice and train supervised probes on the output of the frozen P-JEPA encoder. This separates representation quality from downstream model capacity: if a lightweight probe performs well, the task-relevant information is already directly accessible from the learned feature space.

%\vspace{-0.3cm}
\textbf{Linear Probing.}
\label{par:lin_probe}
Linear probing is our primary diagnostic for representation quality. Since the probe is only an affine classifier, $\boldsymbol{\ell}_i=W\mathbf{x}_i+b$ operating on individual frames, it has no capacity to learn temporal structure on its own. Strong performance therefore suggests that P-JEPA organizes the feature space such that procedural distinctions are close to linearly separable, e.g., visually similar segments can be distinguished by
their position and context within the take.

%\vspace{-0.3cm}
\textbf{Temporal Probing.}
\label{par:TCN}
Since linear probing in an inherently temporal setting can only achieve limited performance we also explore P-JEPAs usefulness on downstream temporal models. We select LTContext \cite{bahrami2023much} and FACT \cite{lu2024fact} as state of the art temporal action segmentation models on Assembly101 \cite{sener2022assembly101} and EgoProceL \cite{bansal2022egoprocel} respectively. 

Both LTContext and FACT are non-causal - they can see future context to make predictions about current frames - so we define two more probes to experiment with online streaming inference. The first is a simple self-attention classifier ingesting a limited past context window to make predictions about current frames and the second is a LTContext version modified to be clip-causal (\textit{LT-Causal}). 
For further technical details we refer the reader to the respective publications and provide implementation details in the supplementary.
\section{Experiments and Results}

%\subsection{Experimental Setup}

\textbf{Implementation Details.}
The total token count is relatively small for each dataset (EgoExo4D: $600k$, EgoProceL: $1.3M$, Assembly101: $300k$), so we use a shallow P-JEPA model with a 4-Layer encoder and 2-Layer predictor for all experiments. The target encoder is an EMA copy of the student with momentum $0.999$. We train for 1000 epochs with learning rate $10^{-4}$, step decay by $0.1$, and random masking of $60$/$80\%$ of tokens, depending on the dataset. We run probing every 20 epochs and save the P-JEPA checkpoint for the best performing probe. We use a frame rate of 4 fps across all datasets.
%\vspace{-0.3cm}

\textbf{Datasets and Metrics.}
\textbf{EgoExo4D}~\cite{grauman2024ego} uses the egocentric fine-grained action classification split with 14,326/4,517 train/validation clips from 648/180 takes and 278 classes; clips average $11.2$s and the longest training take spans $34.5$ minutes. For \textbf{EgoProceL}~\cite{bansal2022egoprocel}, we follow FACT~\cite{lu2024fact}: all reported results use their extracted I3D features, label mapping, and 731/183 train/test split. The dataset contains 914 videos, 117.9 hours, and 19,943 segments over 115 foreground steps plus background; EgoProceL has a large class imbalance, with background accounting for $50.2\%$ of segments and $61.2\%$ of duration. Its longest training video is $36.6$ minutes. For \textbf{Assembly101}~\cite{sener2022assembly101}, we use only fixed RGB view \texttt{C10119} ($v4$) with the pre-extracted TSM features, based on the original papers observation that overhead views $v4$ and $v1$ have the least amount of occlusions. This single-view setting has 393/120 train/validation take streams, 62.5 hours, 6,409 coarse action segments, and a longest training take of $30.8$ minutes.
%\vspace{-0.5cm}
%\textbf{Evaluation metrics.}
As evaluation \textbf{metrics}, for EgoExo4D action classification, we report top-1 accuracy. For temporal action segmentation on EgoProceL and Assembly101, we report frame accuracy \textbf{Acc}, normalized \textbf{Edit Score} over the predicted action sequence, and segmental \textbf{F1@10/25/50}, where the thresholds denote the required temporal IoU between predicted and ground-truth action segments. On EgoProceL, \textbf{Acc$_\text{bg}$} includes background frames.

\subsection{Assembly101}

\begin{table}[h]
\vspace{-0.3cm}
\centering
\caption{\textbf{Assembly101}. We compare P-JEPA's behavior using mean-pool and action-pool V-JEPA2.1 features and TSM features under linear and LTContext probes. P-JEPA can learn procedural structure regardless of input feature source and improves current state of the art LTContext over all metrics.}
\label{tab:assembly101_main}
\small
\renewcommand{\arraystretch}{1.08}

\begin{tabular}{lll>{\columncolor[HTML]{EEF3FF}}cccc}
\toprule
& \multicolumn{3}{c}{\textbf{Representation}} 
& \multicolumn{3}{c}{\textbf{Metrics}} \\
\cmidrule(l){2-4} \cmidrule(lr){5-7}
\textbf{Features} & \textbf{Probe} & \textbf{Pooling} & \textbf{P-JEPA}
& \textbf{Acc.} & \textbf{Edit} & \textbf{F1@0.10 / 0.25 / 0.50} \\
\midrule

\multirow{9.3}{*}{V-JEPA2.1}
& \multirow{4}{*}{Linear} 
    & \multirow{2}{*}{Mean}   &         & 16.1 & 3.3  & 2.7 / 1.9 / 1.1  \\
&  &                          & \checkmark & \textbf{29.4} & \textbf{18.6} & \textbf{16.0} / \textbf{13.5} / \textbf{10.3} \\
\cmidrule(lr){3-7}
&  & \multirow{2}{*}{Action} &          & 34.0 & 18.5 & 17.8 / 15.5 / 10.2 \\
&  &                          & \checkmark & \textbf{40.0} & \textbf{27.8} & \textbf{26.0 / 23.0 / 16.9} \\
\cmidrule(lr){2-7}
& \multirow{4.4}{*}{LTContext} 
    & \multirow{2}{*}{Mean}   &          & 30.6 & 26.2 & 26.0 / 21.7 / 14.3 \\
&  &                          & \checkmark & \textbf{33.8} & \textbf{27.6} & \textbf{29.3} / \textbf{25.0} / \textbf{18.7} \\
\cmidrule(lr){3-7}
&  & \multirow{2}{*}{Action} &          & 41.6 & 32.7 & 34.4 / 30.1 / 21.9 \\
&  &                          & \checkmark & \textbf{43.0} & \textbf{34.2} & \textbf{35.9 / 31.5 / 22.3} \\

\midrule

\multirow{4}{*}{TSM}
& \multirow{2}{*}{Linear}
    & \multirow{2}{*}{--} &          & 23.5 & 3.0  & 2.4 / 1.1 / 0.4 \\
&  &                      & \checkmark & \textbf{40.1} & \textbf{15.0} & \textbf{15.2 / 13.1 / 9.8} \\
\cmidrule(lr){2-7}
& \multirow{2}{*}{LTContext}
    & \multirow{2}{*}{--} &          & 40.2 & 32.2 & 34.3 / 30.5 / 22.4 \\
&  &                      & \checkmark & \textbf{42.9} & \textbf{35.9} & \textbf{37.3 / 34.0 / 26.4} \\

\bottomrule
\end{tabular}

\end{table}

\begin{figure}[t]
    \centering
    \vspace{-0.5cm}
    \includegraphics[width=1.0\linewidth]{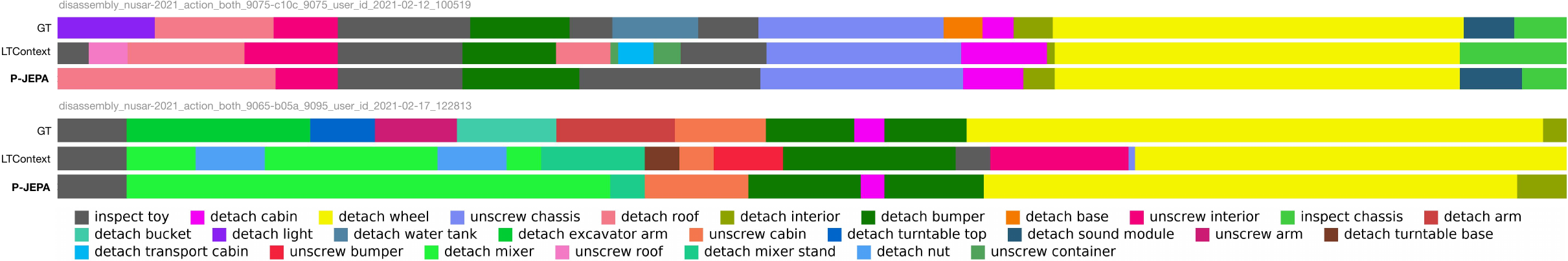}
    \caption{\textbf{Assembly101 Qualitative.} LTContext used with TSM and P-JEPA enhanced TSM features. LTContext can correctly segment \textit{detach sound module} and \textit{detach cabin} when using P-JEPA.}
    \label{fig:quali_assembly}
    \vspace{-0.4cm}
\end{figure}

To investigate P-JEPA's behavior when using different input representations, we experiment with three frame-wise representations on the Assembly101 dataset, \textit{mean-pool} and \textit{action-pool} V-JEPA2.1 features, and the original TSM features that were released with the dataset. Similar to our \textit{action-pool} features, the TSM features were fine-tuned on the dataset to improve domain alignment (see the Supplementary Material of Assembly101 \cite{sener2022assembly101}). 
P-JEPA improves performance over all experiments, bringing the largest linear probing gain for \textit{mean-pool} and TSM features, where the linear probes frame accuracy even matches the temporal LTContext model. 
\textit{Action-pooling} seems to bring large improvements in temporal awareness of the baseline features, even though it only ever sees a limited window of frames, and enhancing them with P-JEPA's long-range procedural knowledge improves further. 
The temporal LTContext probe can utilize P-JEPA features in all settings to improve, although \textit{mean-pool} lags behind, likely due to its feature-agnostic dimensionality reduction and loss of spatial information. LTContext with TSM + P-JEPA features (bottom row) then sets a new state of the art on Assembly101 (in our single view setting), achieving consistent gains over all metrics. This improved performance can also be observed in Figure \ref{fig:quali_assembly}, where P-JEPA features help the model to correctly identify \textit{detach cabin} within the \textit{detach bumper} class and generally aligns action boundaries better.

%We experiment with three main feature sources for P-JEPA on the Assembly101 dataset \cite{sener2022assembly101}. The first set is using V-JEPA2.1 with both \textit{mean-pooling} and \textit{action-pooling} and reported in Table \ref{tab:assembly101_main}. P-JEPA improves action segmentation performance over both settings and probes, with the largest gain for the mean-pooled features. 
%The action-pooled features show much stronger performance under linear and temporal probing, with linear probing showing a much increased temporal awareness of the features compared to mean-pooling. LTContext and action-pooled features perform roughly on par with raw TSM features. 

%The second set of features are the pre-extracted TSM features provided by the dataset. P-JEPA also showcases a significant improvement, with linear probing achieving almost the same frame-wise accuracy as the temporal LTContext model. LTContext with P-JEPA features then sets a new state of the art on Assembly101 (in our single view setting), achieving consistent gains over all metrics. This improved performance can also be observed in Figure \ref{fig:quali_assembly}, where P-JEPA features help the model to correctly identify \textit{detach cabin} within the \textit{detach bumper} class and generally aligns action boundaries better. 

\subsubsection{Ablations}

We ablate our clip-causal design in Table \ref{tab:ablation_temporal_attention} using Assembly101 with TSM features and linear probing, which directly measures procedural/temporal awareness of the features themselves. 
Token-causal enforces a strict lower diagonal attention mask, with a single token only being able to attend to past tokens. This configuration achieves only a small gain over raw TSM features. Bidirectional attention performs better, but even with access to future knowledge, it cannot improve over clip-causal attention, which performs the best. 

\begin{table}[h]
\vspace{-0.2cm}
\centering
\caption{\textbf{Attention Mask Ablation.} Effect of the temporal attention mask in P-JEPA on Assembly101 using TSM features and linear probing.}
\label{tab:ablation_temporal_attention}
\begin{tabular}{lccc}
\toprule
\textbf{Attention} 
& \textbf{Acc.} 
& \textbf{Edit} 
& \textbf{F1@0.10 / 0.25 / 0.50} \\
\midrule
Token-causal      & 35.6 & 7.6  & 7.0 / 5.6 / 1.0 \\
Bidirectional     & 37.6 & 12.2 & 12.2 / 9.8 / 7.2 \\
Clip-causal       & \textbf{40.1} & \textbf{15.0} & \textbf{15.2} / \textbf{13.1} / \textbf{9.8} \\
\bottomrule
\end{tabular}
\vspace{-0.4cm}
\end{table}

\subsection{EgoProcel}

\begin{table}[h]
\vspace{-2mm}
\centering
\small
\caption{\textbf{Streaming Inference.} All experiments use I3D features and streaming inference - only the current segment and its past are visible to the model. LT-Causal is a version of LTContext \cite{bahrami2023much} modified to be clip-causal. P-JEPA is better for all architectures and helps FACT degrade less.}
\label{tab:egoprocel_streaming}
%\scriptsize
%\setlength{\tabcolsep}{4pt}
%\renewcommand{\arraystretch}{1.05}
\begin{tabular}{l >{\columncolor[HTML]{EEF3FF}} c c c c c}
\toprule
\textbf{Model} & \textbf{P-JEPA} & \textbf{Acc$_\text{bg}$} & \textbf{Acc} & \textbf{Edit} & \textbf{F1@0.10 / 0.25 / 0.50} \\
\midrule

\multirow{2}{*}{FACT}
&          & 69.2 & 19.2 & 18.1 & 14.8 / 13.8 / 10.9 \\
& \checkmark & 76.8 & 39.1 & 28.8 & 26.9 / 25.9 / 24.1 \\

\cmidrule(lr){1-6}

\multirow{2}{*}{LT-Causal}
&          & 78.6 & 55.3 & 40.2 & 35.0 / 32.0 / 25.7 \\
& \checkmark & \textbf{80.8} & \textbf{58.8} & \textbf{48.9} & \textbf{42.1 / 40.3 / 34.4} \\

\bottomrule
\end{tabular}
\vspace{-1mm}
\begin{comment}
\begin{tabular}{l c c c c}
\toprule
\textbf{Model} & \textbf{Acc$_\text{bg}$} & \textbf{Acc} & \textbf{Edit} & \textbf{F1@0.10 / 0.25 / 0.50} \\
\midrule
FACT              & 69.2 & 19.2 & 18.1 & 14.8 / 13.8 / 10.9 \\
P-JEPA + FACT (Ours)           & 76.8 & 39.1 & 28.8 & 26.9 / 25.9 / 24.1 \\
LT-Causal         & 78.6 & 55.3 & 40.2 & 35.0 / 32.0 / 25.7 \\
P-JEPA + LT-Causal (Ours)      & \textbf{80.8} & \textbf{58.8} & \textbf{48.9} & \textbf{42.1 / 40.3 / 34.4} \\
\bottomrule
\end{tabular}
%\vspace{-2mm}
\end{comment}
\end{table}

We perform two experiments on EgoProceL \cite{bansal2022egoprocel} using only the I3D features provided by FACT \cite{lu2024fact}: online inference (Table \ref{tab:egoprocel_streaming}) and a sparse label regime (Table \ref{tab:sparsity_i3d_fact}). Online inference is critical for real-time systems, where predictions must be issued as inputs arrive rather than after the full context is observed. To simulate this setting, we perform inference in a streaming fashion, where the model's visible context is extended one segment at a time: at timestep $t$, the model sees only the partial sequence $s_{1:t}$. We evaluate two temporal models on this task, the current state of the art on EgoProceL, FACT (non-causal), and LT-Causal, a version of LTContext we modify to be causal. Both are trained in an offline manner, being able to see entire takes. 
FACT (I3D), which learns latent-action descriptors from the full video, degrades significantly in this setting. FACT (P-JEPA), on the other hand, does not degrade as much, with almost 20 percentage points better Acc. LT-Causal alone performs better than FACT, but also receives a large boost when using P-JEPA features, going from 40.2 to 48.9 Edit Score, indicating it can model the procedural structure better when using P-JEPA features. 

\begin{table}[h]
\vspace{-0.5cm}
\centering
\caption{\textbf{Data Sparsity.} Effect of training-data sparsity on EgoProceL using \textbf{FACT} with I3D and P-JEPA features. FACT + P-JEPA shows large improvements when labeled data is sparse.}
\label{tab:sparsity_i3d_fact}
\small
\setlength{\tabcolsep}{5pt}
\begin{tabular}{lcccccccc}
\toprule
%\multirow{2}{*}{\textbf{Data}}
& \multicolumn{4}{c}{\textbf{I3D}}
& \multicolumn{4}{c}{\textbf{P-JEPA (Ours)}} \\
\cmidrule(lr){2-5} \cmidrule(l){6-9}
\textbf{Data} & \textbf{Acc$_{\mathrm{bg}}$}
& \textbf{Acc.}
& \textbf{Edit}
& \textbf{F1@0.10/0.25/0.50}
& \textbf{Acc$_{\mathrm{bg}}$}
& \textbf{Acc.}
& \textbf{Edit}
& \textbf{F1@0.10/0.25/0.50} \\
\midrule
15\%
& 76.1 & \textbf{50.8} & 41.8 & 41.5 / 37.8 / 26.3
& \textbf{78.9} & 50.1 & \textbf{52.1} & \textbf{48.6} / \textbf{45.6} / \textbf{36.9} \\

33\%
& 80.7 & 61.4 & 53.9 & 52.3 / 49.5 / 38.5
& \textbf{84.5} & \textbf{66.1} & \textbf{65.2} & \textbf{60.1} / \textbf{57.9} / \textbf{52.2} \\

50\%
& 83.7 & 67.5 & 64.6 & 62.5 / 60.0 / 48.1
& \textbf{88.0} & \textbf{74.4} & \textbf{72.7} & \textbf{67.3} / \textbf{65.3} / \textbf{59.8} \\

100\%
& 89.5 & \textbf{81.8} & \textbf{83.7} & \textbf{76.5} / \textbf{74.5} / 68.1
& \textbf{90.9} & 81.2 & 81.6 & 75.0 / 73.5 / \textbf{69.8} \\
\bottomrule
\end{tabular}
%\vspace{-0.2cm}
\end{table}

The second experiment on EgoProceL is about label sparsity (Table \ref{tab:sparsity_i3d_fact}). Here, we want to test the performance of FACT when less than the full dataset is available for supervised training. We therefore subsample the training set to $15\%, 33\%, 50\%$, while maintaining full class coverage. The method is then trained on the limited training set and evaluated on the untrimmed validation set. All experiments here use the same hyperparameters. 
P-JEPA features result in almost an entire step up in performance, where FACT (P-JEPA) trained on only $33\%$ of data achieves results on par with or better than FACT (I3D) trained on $50\%$ of data. Only at $100\%$ of the data does FACT (I3D) regain a small advantage, although FACT (P-JEPA) is still better in the hardest F1 metric by 1.7 points.

\begin{figure}[h]
    \centering
    %\vspace{-0.3cm}
    \includegraphics[width=1.0\linewidth]{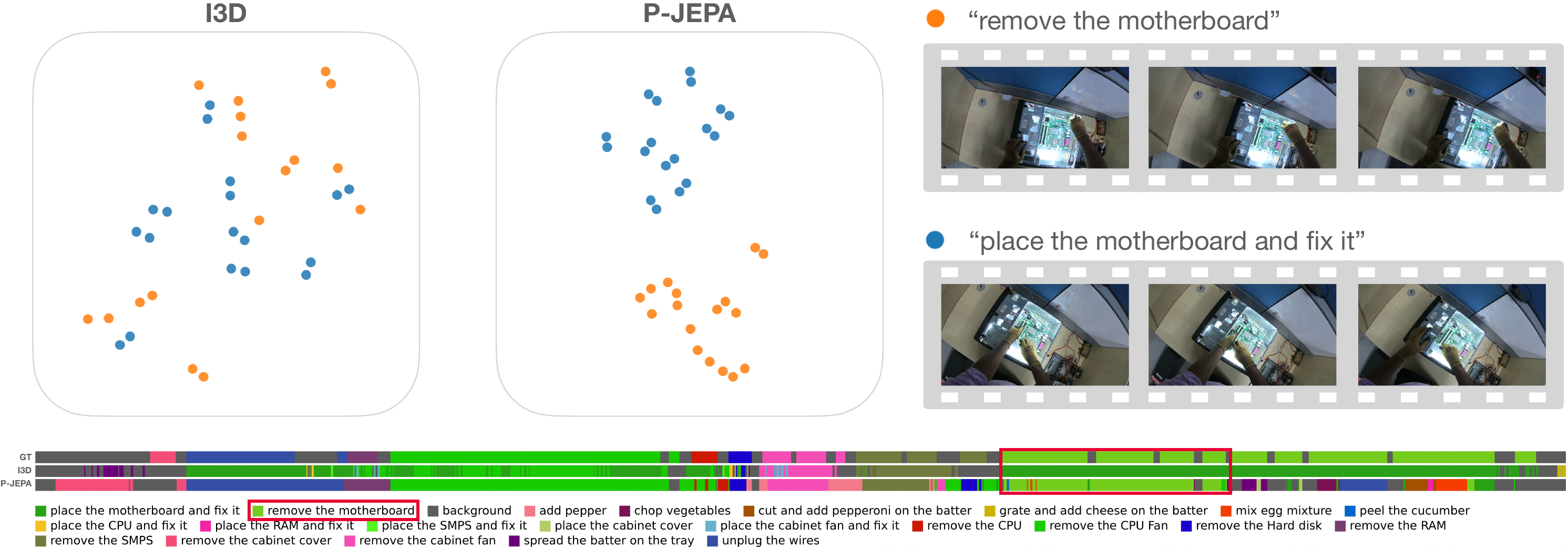}
    \caption{\textbf{t-SNE} of original I3D and P-JEPA features for two visually similar classes from EgoProceL validation set. P-JEPA features form two separate clusters. Full take linear probing results below confirm improved linear separability, with I3D features failing on the \textit{remove the motherboard} class.}
    \label{fig:tsne_motherboard}
    \vspace{-0.6cm}
\end{figure}

In Figure \ref{fig:tsne_motherboard} we take a closer look at two visually extremely similar classes from the EgoProceL validation set, \textit{place the motherboard and fix it} and \textit{remove the motherboard}, where both actions require (un-)screwing the motherboard. An example of both actions is illustrated on the right of the figure and can only really be distinguished from context. Has the motherboard been recently placed in the case, or are we currently in the process of removing it? 
P-JEPA answers this question in feature space, where its t-SNE exhibits two distinctly separate clusters compared to the raw I3D features. 
Additionally, we illustrate this case for an entire takes action segmentation performance below using linear probing, where P-JEPA lets the probe distinguish the \textit{remove the motherboard} class at all.

\subsection{EgoExo4D - Fine-grained Action Classification}

\begin{table}[h]
\vspace{-0.2cm}
\caption{\textbf{EgoExo4D Action Classification}. Results on the official benchmark. \textit{egoX} indicates training with ego+exo data.}
\centering
\small
\begin{tabular}{llcc}
\toprule
\textbf{Model} & \textbf{Train Data} & \textbf{LLM} & \textbf{Ego Acc. (\%)} \\
\midrule
TimeSformer \cite{bertasius2021space}              & ego  & no  & 35.13 \\
EgoVLPv2 \cite{pramanick2023egovlpv2}              & egoX & no  & 39.10 \\
Viewpoint Distillation                             & egoX & no  & 38.19 \\
View-Invariant Enc. \cite{oord2018representation} & egoX & no  & 40.34 \\
\midrule
VideoLLM-MoD \cite{wu2024videollm}                 & ego  & yes & 42.62 \\
ProVideLLM-8B/11 \cite{chatterjee2025streaming}    & ego  & yes & 44.36 \\
\midrule
\cellcolor[HTML]{EEF3FF}
\textbf{P-JEPA} + Streaming Probe \textbf{(ours)}           & ego  & no  & \textbf{56.85} \\
\bottomrule
\end{tabular}
\label{tab:benchmark}
\vspace{-0.3cm}
\end{table}

To provide a comparison with LLM-based video understanding, we also experiment on the EgoExo4D fine-grained action classification benchmark. Here, we use only \textit{action-pooled} V-JEPA2.1 features and a streaming classifier, which ingests a limited context window to make predictions about the current action. This is in line with prior works like ProVideLLM \cite{chatterjee2025streaming} and the overall benchmark design, which prohibits future information leakage. P-JEPA achieves state-of-the-art results with a large increase over LLM baselines, while having only 1.2B total parameters, almost an order of magnitude less than typical VideoLLMs. Our entire model stack, including V-JEPA2.1 encoder, pooler, P-JEPA encoder, and streaming classifier, also runs at around 20fps (no decoding overhead, inference only) on a consumer grade RTX 3090 without major optimization, making it real-time capable. 

\section{Limitations \& Conclusion}
\label{sec:conclusion}
\textbf{Limitations.}
P-JEPA learns the most immediately useful procedural representations with clip-causal attention, which requires segment boundaries. These are not guaranteed to be present in a real-world setting or when running inference on live video capture. Exploring a simple fixed-window block-causal attention mechanism could remove the reliance on this mechanism, but we leave this to future work. 
P-JEPA also requires frame-aligned token representations that encode knowledge about the underlying low-level actions (e.g., interactions, motion, direction). V-JEPA2.1 can provide such features for most datasets, but its mean-pool variant performed the worst in our experiments, while action-pooling requires labeled data to train. Combining this with video token compression \cite{lee2024video} is a promising future direction.

\textbf{Conclusion.}
We presented P-JEPA, a procedural representation learning model that discovers procedural information from dense frame-aligned features of over 30-minute-long videos. It achieves this through a masked latent prediction objective, without relying on narrations, transcripts, or procedural knowledge graphs.
P-JEPA representations enhance temporal understanding tasks, and, combined with strong temporal models, achieve state-of-the-art results while running in real-time and requiring significantly fewer parameters than current VideoLLMs.

{
    \small
    \bibliographystyle{plainnat}
    \bibliography{main}
}

%%%%%%%%%%%%%%%%%%%%%%%%%%%%%%%%%%%%%%%%%%%%%%%%%%%%%%%%%%%%
\newpage
\appendix

\section{Supplementary Material}

\begin{figure}[h]
    \centering
    \includegraphics[width=0.7\linewidth]{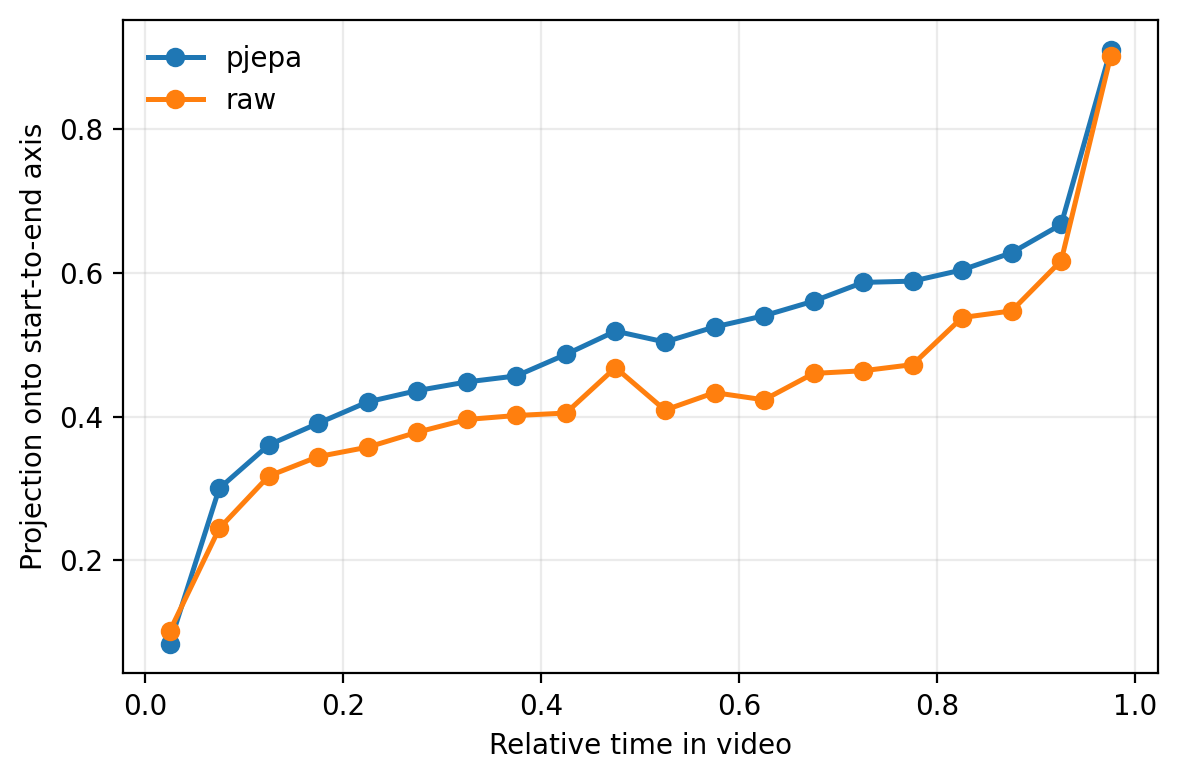}
    \caption{
Average feature-space progress over relative video time.
For each video, segment features are projected onto the video's own start--end feature direction and averaged across relative-time bins.
A diagonal increasing curve indicates that representation change is aligned with temporal progress.
P-JEPA exhibits a more monotonic progression than raw features, matching the progress metrics in Table~\ref{tab:egoprocel_feature_trajectory}.
}

    \label{fig:egoprocel_progress_visual}
\end{figure}

\begin{figure}[h]
    \centering
    \includegraphics[width=1.0\linewidth]{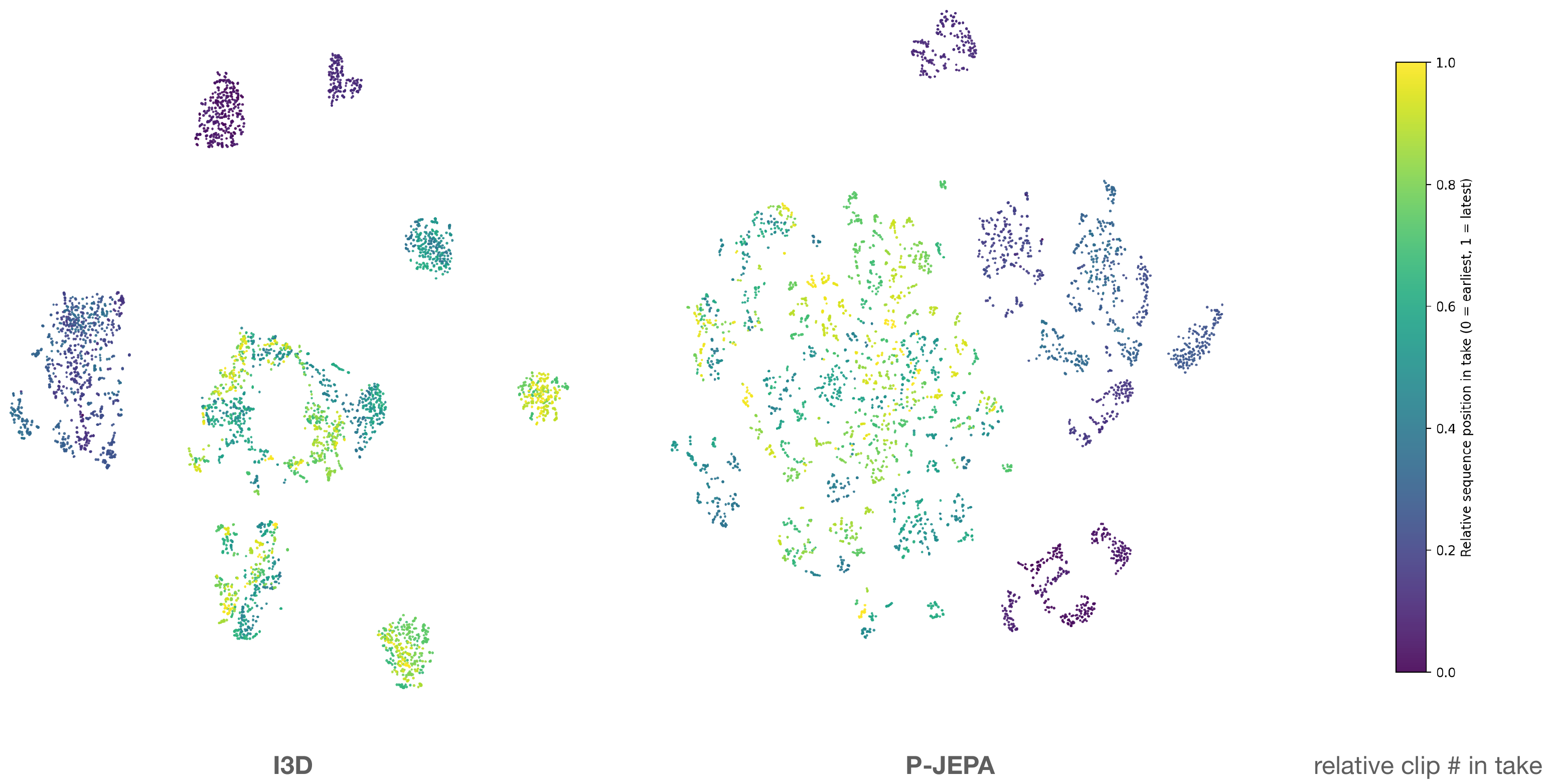}
    \caption{t-SNE of original and P-JEPA enhanced I3D features from the EgoProceL validation split, colored by relative occurrence during the video. P-JEPA rearranges representations around their occurrence inside a procedure.}
    \label{fig:tsne_both}
\end{figure}

\begin{figure}[h]
    \centering
    \includegraphics[width=1.0\linewidth]{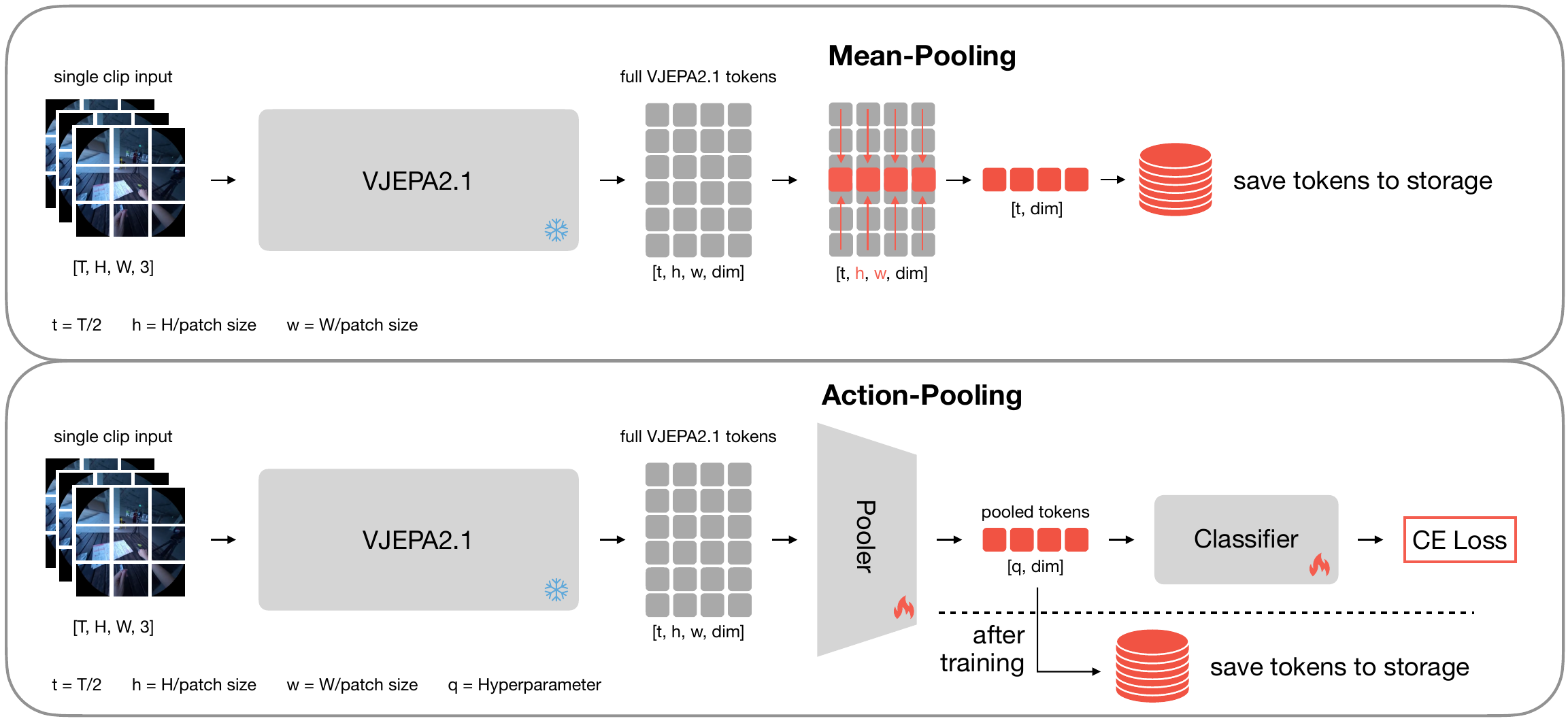}
    \caption{Overview of Pooling frameworks.}
    \label{fig:pooling_framework}
\end{figure}

In this supplementary material we first discuss the broader societal impact of our work in Section \ref{sec:impact}, then detail the compute resources used in Section \ref{sec:compute}. 
In Section \ref{sec:suppl_probe_details} we provide further information on the types of probes used and their hyperparameters. 
Figure \ref{fig:pooling_framework} illustrates the V-JEPA2.1 pooling architectures that we used in the main paper and which were discussed in Section \ref{sec:features}. 
In the end we provide more insight into the representations that P-JEPA learns, by first measuring similarity metrics along the video path in feature space in Section \ref{sec:feature_analysis} and outlining our findings in Table \ref{tab:egoprocel_feature_trajectory} and Figure \ref{fig:egoprocel_progress_visual}. Then we perform a similar analysis in the reduced dimensionality space of the t-SNE as shown in Figure \ref{fig:tsne_both}, where the path metrics were computed in this reduced space (Table \ref{tab:egoprocel_tsne_trajectory}). 

\subsection{Broader Impact}
\label{sec:impact}

Embodied AI and personalized assistants have a large upside, promising to take over menial tasks and increase productivity, making life easier for the majority of people. The transition to fully autonomous human-like systems, however, is not without drawbacks, where in the short-term human employment might decrease. 
Training models to achieve this vision also consumes large amounts of energy, which has an impact on the global climate. 

Our work provides one stepping stone on the path towards this goal, but its training is rather small scale compared to other works and there are still many other hurdles to overcome before autonomous embodied AI systems are in widespread use. 

\subsection{Compute Resources}
\label{sec:compute}

We train P-JEPA models on one Nvidia H200 GPU (141GB), the action-pooling models were trained on one H100 (96GB) and some probing experiments were done locally on a consumer grade RTX 3090 (24GB). Inference speed was measured on the same 3090 as well, using a dummy setup where we stitch together all components that are usually separated for easier dataloading. 
We use python 3.11/3.12, PyTorch 2.6.0 and CUDA 12.6 for all experiments. 
Training a P-JEPA model takes between 8-12 hours depending on the dataset, training time of the probes depends on the type used. Epoch counts are outlined in the following hyperparameter tables. 

\subsection{Further Feature-Space Analysis}
\label{sec:feature_analysis}

\begin{table}[h]
\centering
\caption{
Feature-space temporal trajectory diagnostics on EgoProceL validation videos.
Values report medians across videos. Metrics are computed on L2-normalized pooled segment features, before PCA or t-SNE.
}
\small
\begin{tabular}{lcc}
\toprule
Metric & Raw & P-JEPA \\
\midrule
Trajectory straightness $\uparrow$ & 0.0589 & \textbf{0.0839} \\
Time-progress Spearman $\uparrow$ & 0.6078 & \textbf{0.9351} \\
Progress efficiency $\uparrow$ & 0.3512 & \textbf{0.7685} \\
Monotonic progress fraction $\uparrow$ & 0.5870 & \textbf{0.7059} \\
Adjacent temporal NN fraction $\uparrow$ & 0.4737 & \textbf{0.8000} \\
Adjacent cosine distance $\downarrow$ & \textbf{0.0339} & 0.3376 \\
Start--end displacement $\uparrow$ & 0.2632 & \textbf{1.1355} \\
\bottomrule
\end{tabular}
\label{tab:egoprocel_feature_trajectory}
\end{table}

\paragraph{Feature-space temporal trajectory diagnostics.}
To analyze whether learned representations evolve coherently over time, we compute trajectory statistics directly in the original feature space, before PCA or t-SNE. For each video, we order its temporal segments by timestamp and represent each segment by its L2-normalized pooled feature vector. This gives a sequence of feature vectors $(z_1, \ldots, z_T)$.

We report several diagnostics. \emph{Adjacent cosine distance} measures local change between consecutive segments, $1 - z_t^\top z_{t+1}$; lower values indicate locally smoother representations. \emph{Start--end displacement} measures the Euclidean distance between the first and last segment, $\|z_T - z_1\|_2$, capturing how much the representation changes over the full video. \emph{Trajectory straightness} is the ratio between start--end displacement and total path length,
\[
\frac{\|z_T - z_1\|_2}{\sum_{t=1}^{T-1} \|z_{t+1} - z_t\|_2},
\]
where higher values indicate that feature changes are more consistently directed from the beginning toward the end of the video.

We also measure whether representation change aligns with temporal progress. For each video, we define a start--end axis $u = z_T - z_1$ and project each segment onto this axis. \emph{Time-progress Spearman} is the Spearman correlation between segment time index and this projected progress; higher values indicate that later segments tend to lie farther along the video-specific start--end direction. \emph{Progress efficiency} measures how directly the projected trajectory moves from start to end, penalizing backtracking along this axis. \emph{Monotonic progress fraction} is the fraction of consecutive segment transitions whose projected progress is non-decreasing. Finally, \emph{adjacent temporal nearest-neighbor fraction} measures how often a segment's nearest neighbor in feature space is one of its immediate temporal neighbors, indicating whether local feature neighborhoods respect temporal adjacency.

These diagnostics should not be interpreted as downstream task performance. Instead, they provide a feature-space characterization of temporal organization. In particular, raw features are highly locally smooth, but this can reflect limited representation change over time. P-JEPA features move more substantially through feature space while showing stronger alignment between representation change and temporal progress.

\paragraph{Progress profile visualization.}
To visualize the temporal progress metrics, we project each video's segment features onto that video's own start--end feature direction. For a video with ordered segment features $(z_1,\ldots,z_T)$, we define the start--end direction as
\[
u = z_T - z_1.
\]
Each segment feature $z_t$ is then projected onto this direction:
\[
p_t = \frac{(z_t - z_1)^\top u}{\|u\|_2^2}.
\]
Under this normalization, $p_1 = 0$ and $p_T = 1$. Thus, $p_t$ measures how far segment $t$ has progressed from the video's initial representation toward its final representation. We then bin segments by their relative timestamp in the video and average $p_t$ across videos within each bin.

An idealized representation whose change is perfectly aligned with video time would produce an approximately diagonal curve from $(0,0)$ to $(1,1)$: as relative time increases, projected feature progress would increase at the same rate. Flat regions indicate little representational progress, while downward fluctuations indicate backtracking along the video's start--end feature direction.

This visualization corresponds directly to the progress-based metrics in Table~\ref{tab:egoprocel_feature_trajectory}. \emph{Time-progress Spearman} measures, for each video, whether later segments have larger projected progress values $p_t$. \emph{Progress efficiency} measures how directly the projected trajectory moves from $p_1=0$ to $p_T=1$, penalizing backtracking. \emph{Monotonic progress fraction} measures the fraction of consecutive segment pairs for which projected progress does not decrease. The P-JEPA curve is closer to a monotonic diagonal trend than the raw-feature curve, consistent with its higher time-progress Spearman, progress efficiency, and monotonic progress fraction.

\begin{table}[h]
\caption{
Trajectory diagnostics measured in the 2D t-SNE embedding of EgoProceL validation videos.
Values report medians across videos. These metrics characterize the geometry of the visualization and should not be interpreted as distances in the original feature space.
}
\centering
\small
\begin{tabular}{lcc}
\toprule
Metric & Raw & P-JEPA \\
\midrule
Trajectory straightness $\uparrow$ & 0.0282 & \textbf{0.2141} \\
Median step length $\downarrow$ & 1.2410 & \textbf{0.5858} \\
90th-percentile step length $\downarrow$ & 5.4648 & \textbf{3.1594} \\
Maximum step length $\downarrow$ & \textbf{7.7705} & 10.0568 \\
Total path length $\downarrow$ & 37.5821 & \textbf{33.3859} \\
Start--end displacement $\uparrow$ & 1.3174 & \textbf{6.2600} \\
\bottomrule
\end{tabular}
\label{tab:egoprocel_tsne_trajectory}
\end{table}

\paragraph{t-SNE trajectory diagnostics.}
In addition to feature-space diagnostics, we compute analogous trajectory statistics on the 2D t-SNE embeddings used for visualization. For each video, temporal segments are ordered by timestamp and connected in the 2D embedding. We then measure the resulting projected path. \emph{Median step length}, \emph{90th-percentile step length}, and \emph{maximum step length} summarize the distances between consecutive temporal segments in the 2D map. \emph{Total path length} is the sum of all consecutive step lengths. \emph{Start--end displacement} is the Euclidean distance between the first and last segment in the embedding. \emph{Trajectory straightness} is the ratio between start--end displacement and total path length, with larger values indicating a more direct projected path.

These metrics are useful only as diagnostics of the visualization. Since t-SNE preserves local neighborhoods but does not preserve global distances, path lengths, or angles, these measurements should not be interpreted as feature-space geometry. We therefore use them as a qualitative complement to the original feature-space trajectory metrics rather than as primary evidence.

\subsection{Implementation Details}
\label{sec:suppl_probe_details}

\begin{comment}
\subsection{Take-Level P-JEPA}

All active P-JEPA runs use the feature-stream implementation described in
Section~\ref{sec:ssl_procedure}. Input features are padded in batches, but the
model receives the valid-token mask and, for flat continuous streams, the
segment lengths. These lengths define the segment coordinate $c(i)$ and local
coordinate $r(i)$ used by 2D RoPE and by the clip-causal attention mask. The
student and teacher encoders use the same clip-causal mask; the predictor uses
token-causal attention over the flattened stream.

The take-level model uses a 4-layer encoder and a 2-layer predictor. The teacher
is updated as an EMA of the student encoder with momentum $0.999$. We train with
AdamW and weight decay $0.04$. The reported runs use random token masking only,
with mask ratio $0.8$ for EgoExo4D and EgoProceL, and $0.6$ for Assembly101 TSM
runs.
\end{comment}

\subsubsection{Probe Details}

For EgoProceL, all probe results use the official FACT I3D features, label
mapping, and \texttt{split1} train/test split. P-JEPA is trained and evaluated
on those same 2048-d feature streams.

\begin{table}[h]
\centering
\caption{Linear probe configuration.}
\label{tab:supp_linear_probe_details}
\footnotesize
\setlength{\tabcolsep}{4pt}
\renewcommand{\arraystretch}{1.08}
\begin{tabular}{@{}p{0.34\linewidth}p{0.58\linewidth}@{}}
\toprule
\textbf{Setting} & \textbf{Value} \\
\midrule
Input features & Frozen raw or P-JEPA-contextualized features \\
Head & Affine classifier per token/frame \\
Optimizer & AdamW \\
Epochs & 30 \\
Learning rate & $10^{-3}$ \\
Weight decay & $0$ \\
Schedule & MultiStep decay at epochs 5 and 15 with $\gamma=0.1$ \\
\bottomrule
\end{tabular}
\end{table}

\begin{table}[h]
\centering
\caption{LTContext probe configuration for Assembly101 TSM experiments.}
\label{tab:supp_ltcontext_probe_details}
\footnotesize
\setlength{\tabcolsep}{4pt}
\renewcommand{\arraystretch}{1.08}
\begin{tabular}{@{}p{0.34\linewidth}p{0.58\linewidth}@{}}
\toprule
\textbf{Setting} & \textbf{Value} \\
\midrule
Task / dataset & Temporal action segmentation on Assembly101 \\
Feature stream & TSM, with and without P-JEPA contextualization \\
Stages & 4 \\
Layers per stage & 9 \\
Model dimension & 128 \\
Windowed-attention width & 16 \\
Long-term attention stride & 64 \\
Convolution dilation factor & 4 \\
Attention heads & 1 \\
Dropout & 0.2 \\
Channel masking & 0.3 \\
Dimensional reduction & None between stages \\
Loss & Stage-wise cross-entropy plus LTContext temporal smoothing, weight 0.17 and clipping value 16 \\
Optimizer & AdamW, weight decay $0$ \\
Epochs / learning rate & 30 epochs, $2.5\times10^{-4}$ \\
Schedule & Learning-rate drops at epochs 10 and 20 \\
\bottomrule
\end{tabular}
\end{table}

\paragraph{Causal temporal probes.}
The causal LTContext probe keeps the LTContext multi-stage structure but replaces
all temporal operations with prefix-safe counterparts. Dilated convolutions are
left-padded, and both windowed attention and long-term sparse attention use the
segment-causal mask
\[
    A^{\mathrm{LTC}}_{ij}
    =
    \mathbf{1}[i,j\ \mathrm{valid}]
    \mathbf{1}[c(j)\le c(i)] .
\]
Thus a token may use its own segment and previous segments, but never future
segments.

\begin{table}[h]
\centering
\caption{Causal LTContext probe configuration for EgoProceL streaming temporal segmentation.}
\label{tab:supp_causal_ltcontext_probe_details}
\footnotesize
\setlength{\tabcolsep}{4pt}
\renewcommand{\arraystretch}{1.08}
\begin{tabular}{@{}p{0.34\linewidth}p{0.58\linewidth}@{}}
\toprule
\textbf{Setting} & \textbf{Value} \\
\midrule
Task / dataset & EgoProceL temporal action segmentation \\
Feature stream & Official FACT I3D features, with and without P-JEPA contextualization \\
Causality & Current segment and previous segments only; no future-segment access \\
Stages & 3 \\
Layers per stage & 6 \\
Model dimension & 128 \\
Windowed-attention width & 16 \\
Long-term attention stride & 64 \\
Convolution dilation factor & 2 \\
Dimensional reduction & 2 \\
Attention heads & 1 \\
Dropout / channel masking & 0.2 / 0.3 \\
Optimizer & AdamW, weight decay $0$ \\
Epochs / learning rate & 30 epochs, $2.5\times10^{-4}$ \\
Schedule & Learning-rate drops at epochs 15 and 25 \\
\bottomrule
\end{tabular}
\end{table}

\begin{table}[h]
\centering
\caption{FACT probe configuration for EgoProceL temporal action segmentation.}
\label{tab:supp_fact_probe_details}
\footnotesize
\setlength{\tabcolsep}{4pt}
\renewcommand{\arraystretch}{1.08}
\begin{tabular}{@{}p{0.34\linewidth}p{0.58\linewidth}@{}}
\toprule
\textbf{Setting} & \textbf{Value} \\
\midrule
Evaluation setup & Official FACT I3D features, label mapping, and \texttt{split1} train/test split \\
Input features & 2048-d I3D features, with and without P-JEPA contextualization \\
Streaming protocol & For streaming rows, FACT receives complete segments up to the current segment and predictions are kept only for the current segment \\
FACT block string & \texttt{iUUU} \\
Action tokens & 200 \\
Cross-mask ratio & 0.3 \\
Matching weight & 0.9 \\
\texttt{Bi} action block & Action dimension 256, feed-forward dimension 512, 6 action layers, 8 heads \\
\texttt{Bi} frame block & Frame module \texttt{m2}, frame dimension 256, 10 frame layers, hidden dimension 512 \\
Optimizer & Adam \\
Epochs / learning rate & 150 epochs, $10^{-4}$ \\
Gradient clipping & 10 \\
Schedule & Learning-rate drop at epoch 100 \\
\bottomrule
\end{tabular}
\end{table}

\begin{table}[h]
\centering
\caption{Streaming attentive probe configuration.}
\label{tab:supp_streaming_attentive_probe_details}
\footnotesize
\setlength{\tabcolsep}{4pt}
\renewcommand{\arraystretch}{1.08}
\begin{tabular}{@{}p{0.30\linewidth}p{0.30\linewidth}p{0.30\linewidth}@{}}
\toprule
\textbf{Setting} & \textbf{EgoExo4D} & \textbf{EgoProceL} \\
\midrule
Input visibility & Prefix window only & Prefix window only \\
Prediction stitching & Keep predictions for current chunk & Keep predictions for current chunk \\
Stream window $W_{\mathrm{stream}}$ & 320 tokens & 400 tokens \\
Output chunk & 64 tokens & 128 tokens \\
Projection dimension & 1280 & 1408 \\
Transformer depth / heads & 1 layer / 4 heads & 1 layer / 4 heads \\
Position embeddings & Learned absolute positions inside window & Learned absolute positions inside window \\
Dropout & 0.2 & 0.2 \\
Optimizer & AdamW & AdamW \\
Learning rate & $10^{-4}$ & $10^{-4}$ \\
Weight decay & 0.005 & 0.005 \\
Epochs & 30 & 30 \\
Schedule & Drops at epochs 10 and 20 & Drops at epochs 10 and 20 \\
\bottomrule
\end{tabular}
\end{table}

%%%%%%%%%%%%%%%%%%%%%%%%%%%%%%%%%%%%%%%%%%%%%%%%%%%%%%%%%%%%
%\FloatBarrier
%\newpage
%\input{checklist.tex}

\end{document}